\documentclass{llncs}

\usepackage{amsmath}
\usepackage{amssymb}
\usepackage{bm}
\usepackage{array}
\usepackage{xcolor}
\usepackage{multirow}

\usepackage{listings}
\usepackage{tikz}
\usepackage{pgfplots}
\usepackage{subfig}

\usepackage{calc}
\usepackage{filecontents}

\usepackage{url}
\usepackage{caption}
\usepackage{float}

\begin{document}

\title{Labeling Topics with Images using a Neural Network}
\titlerunning{Labeling Topics with Images using a Neural Network}  
%
\author{Nikolaos Aletras\inst{1} \and Arpit Mittal\inst{1}}
%
\authorrunning{Nikolaos Aletras et al.} 
%
%
\institute{Amazon.com, Cambridge, UK\\
\email{\{aletras, mitarpit\}@amazon.com}}

\maketitle              

\begin{abstract}
Topics generated by topic models are usually represented by lists of $t$ terms or alternatively using short phrases or images. The current state-of-the-art work on labeling topics using images selects images by re-ranking a small set of candidates for a given topic. In this paper, we present a more generic method that can estimate the degree of association between any arbitrary pair of an unseen topic and image using a deep neural network. Our method achieves better runtime performance $O(n)$ compared to $O(n^2)$ for the current state-of-the-art method, and is also significantly more accurate.
\end{abstract}
\section{Introduction}\label{sec:introduction}
Topic models \cite{Blei2003} are a popular method for organizing and interpreting large document collections by grouping documents into various thematic subjects (e.g. sports, politics or lifestyle) called topics. Topics are multinomial distributions over a predefined vocabulary whereas documents are represented as probability distributions over topics. Topic models have proven to be an elegant way to build exploratory interfaces (i.e. topic browsers) for visualizing document collections by presenting to the users lists of topics \cite{Chaney2012,Snyder2013,Smith2014} where they select documents of a particular topic of interest. 

A topic is traditionally represented by a list of $t$ terms with the highest probability. In recent works, short phrases \cite{Lau2011,Aletras2014a}, images \cite{Aletras2013a} or summaries \cite{summariesACL16} have been used as alternatives. Particularly, images offer a language independent representation of the topic which can also be complementary to textual labels. The visual representation of a topic has been shown to be as effective as the textual labels on retrieving information using a topic browser while it can be understood quickly by the users \cite{Aletras2014b,Aletras2015}. The task of labeling topics consists of two main components: (1) a candidate generation component where candidate labels are obtained for a given topic (usually using information retrieval techniques and knowledge bases \cite{Lau2011,Aletras2013a}), and (2) a ranking (or label selection) component that scores the candidates according to their relevance to the topic. In the case of labeling topics with images the candidate labels consist of images.

The method presented by \cite{Aletras2013a} generates a graph where the candidate images are its nodes. The edges are weighted with a similarity score between the images that connect. Then, an image is selected by re-ranking the candidates using PageRank. The method is iterative and has a runtime complexity of $O(n^2)$ which makes it infeasible to run over large number of images. Hence, for efficiency the candidate images are selected a priori using an information retrieval engine. Thus the scope of this method gets limited to solving a \emph{local problem} of re-ordering a small set of candidate images for a given topic. Furthermore, its accuracy is limited by the recall of the information retrieval engine. Finally, if new candidates appear, they should be added to the graph, the process of computing pairwise similarities and re-ranking of nodes is repeated.

In this work, we present a more generic method that directly estimates the appropriateness of any arbitrary pair of topic and image. We refer to this method as a \emph{global method} to differentiate it from the localized approach described above. We utilize a Deep Neural Network (DNN) to estimate the suitability of an image for labeling a given topic. DNNs have proven to be effective in various IR and NLP tasks \cite{Collobert2008,Socher2011}. They combine multiple layers that perform non-linear transformations to the data allowing the automatic learning of high-level abstractions.  
At runtime our method computes dot products between various features and the model weights to obtain the relevance score, that gives it an order complexity of $O(n)$. Hence, it is suitable for using it over large image sources such as Flickr\footnote{\url{http://www.flickr.com}}, Getty\footnote{\url{http://www.gettyimages.co.uk}} or ImageNet \cite{deng2009}. The proposed model obtains state-of-the-art results for labeling completely unseen topics with images compared to previous methods and strong baselines.

\section{Model}\label{sec:model}

For a topic $T$ and an image $I$, we want to compute a real value $s \in \mathbb{R}$ that denotes how good the image $I$ is for representing the topic $T$. $T$ consists of ten terms ($t$) with the highest probability for the topic. We denote the visual information of the image as $V$. The image is also associated with text in its caption, $C$.

For the topic $T = \{t_1, t_2,...,t_{10}\}$ and the image caption $C = \{c_1, c_2,...,c_{n}\}$, each term is transformed into a vector $\mathbf{x} \in \mathbb{R}^d$ where $d$ is the dimensionality of the distributed semantic space. We use pre-computed dependency-based word embeddings \cite{Levy2014} whose $d$ is 300. The resulting representations of $T$ and $C$ are the mean vectors of their constituent words, $\mathbf{x_t}$ and $\mathbf{x_c}$ respectively. 

The visual information from the image $V$ is converted into a dense vectorized representation, $\mathbf{x_v}$. That is the output of the publicly available 16-layer VGG-net \cite{simonyan2014} trained over the ImageNet dataset \cite{deng2009}. VGG-net provides a 1000 dimensional vector which is the soft-max classification output of ImageNet classes.

The input to the network is the concatenation of topic, caption and visual vectors. i.e., 
\begin{equation} \label{eq1}
X = [x_t || x_c || x_v]
\end{equation}
This results in a 1600-dimensional input vector.

Then, $X$ is passed through a series of four hidden layers, $H_1,...,H_4$. In this way the network learns a combined representation of topics and images and the non-linear relationships that they share.
\begin{equation} \label{eq2}
h_i = g(W_i^Th_{i-1})
\end{equation}
where $g$ is the rectified linear unit (ReLU) and $h_0 = X$. The output of each hidden layer is regularized using dropout \cite{srivastava2014}. The output size of $H_1, H_2, H_3$ and $H_4$ are set to 256, 128, 64 and 32 nodes respectively. 

The output layer of the network maps the input to a real value $s \in \mathbb{R}$ that denotes how good the image $I$ is for the topic $T$. The network is trained by minimizing the mean absolute error:
\begin{equation} \label{eq3}
error = \frac{1}{n}\sum_{i=1}^{n}\left | W_o^Th_4 - s_g \right |
\end{equation}
where $s_g$ is the ground-truth relevance value. The network is optimized using a standard mini-batch gradient descent method with RMSProp adaptive learning rate algorithm \cite{tieleman2012}.

\section{Experimental Setup}\label{sec:evaluation}

We evaluate our model on the publicly available data set provided by \cite{Aletras2013a}. It consists of 300 topics generated using Wikipedia articles and news articles taken from the New York Times. Each topic is represented by ten terms with the highest probability. They are also associated with 20 candidate image labels and their human ratings between 0 (lowest) and 3 (highest) denoting the appropriateness of these images for the topic. That results into a total of 6K images and their associated textual metadata which are considered as captions. The task is to choose the image with the highest rating from the set of the 20 candidates for a given topic.

The 20 candidate image labels per topic are collected by \cite{Aletras2013a} using an information retrieval engine (Google). Hence most of them are expected to be relevant to the topic. This jeopardizes the training of our supervised model due to the lack of sufficient negative examples. To address this issue we generate extra negative examples. For each topic we sample 
another 20 images from random topics in the training set and assign them a relevance score of 0. These extra images are added into the training data. 

Our evaluation follows prior work \cite{Lau2011,Aletras2013a} using two metrics.  The \textbf{Top-1 average rating} is the average human rating assigned to the top-ranked label proposed by the topic labeling method. This metric provides an indication of the overall quality of the label selected and takes values from 0 (irrelevant) to 3 (relevant). The normalized discounted cumulative gain (\textbf{nDCG}) compares the label ranking proposed by the labeling method to the gold-standard ranking provided by the human annotators \cite{Jarvelin2002,Croft2009}. 

We set the dropout value to 0.2 which randomly sets 20\% of the input units to 0 at each update during the training time. We train the model in a 5-fold cross-validation for 30 epochs and set the batch size for training data to 16. In each fold, data from 240 topics are used for training which results into 9,600 examples (20 original, 20 negative candidates per topic). The rest completely unseen 60 topics are used for testing which results into 1,200 test examples (note that we do not add negative examples in the test data).

\section{Results and Discussion}\label{sec:results}

\begin{table*}[!t]
\footnotesize
\renewcommand{\arraystretch}{1.0}
\begin{center}
    \begin{tabular}{|l|c|c|c|c|}
    \hline
	\textbf{Model}&\textbf{Top-1 aver. rating}&\textbf{nDCG-1}&\textbf{nDCG-3}&\textbf{nDCG-5}   \\ 
	\hline
    Global PPR \cite{Aletras2013a}	&1.89						&0.71	&0.74	&0.75	\\
    Local PPR \cite{Aletras2013a} 	&2.00						&0.74	&0.75	&0.76  	\\ 
    WSABIE \cite{Weston2010}		&1.87						&0.65	&0.68	&0.70	\\
    LR (Topic+Caption+VGG) 			&1.91						&0.71	&0.74	&0.75	\\
    SVM (Topic+Caption+VGG) 		&1.94						&0.72	&0.75	&0.76	\\
    \hline
    DNN (Topic+Caption)				&1.94						&0.73	&0.75	&0.76	\\
    DNN (Topic+VGG)					&$2.04^{\ddagger\ast}$		&0.76	&0.79	&0.80	\\
    DNN (Topic+Caption+VGG)		&$\mathbf{2.12}^{\dagger\ddagger\ast}$	&$\mathbf{0.79}$	&$\mathbf{0.80}$	&$\mathbf{0.81}$	\\
    \hline
    Human Perf. \cite{Aletras2013a}			&2.24						&-	&-	&-	\\
    \hline
    \end{tabular} 
\caption{Results obtained for the various topic labeling methods. $\dagger$, $\ddagger$ and $\ast$ denote statistically significant difference to Local PPR, Global PRR and WSABIE respectively (paired t-test, $p<0.01$).}\label{table:results}    
\end{center}
\end{table*} 

We compare our approach to the state-of-the-art method that uses Personalized PageRank \cite{Aletras2013a} to re-rank image candidates ({\bf Local PPR}) and an adapted version that computes the PageRank scores of all the available images in the test set ({\bf Global PPR}). We also test other baselines methods: (1) a relevant approach originally proposed for image annotation that learns a joint model of text and image features ({\bf WSABIE}) \cite{Weston2010}, (2) linear regression and SVM models that use the concatenation of the topic, the caption and the image vectors as input, {\bf LR (Topic+Caption+VGG)} and {\bf SVM (Topic+Caption+VGG)} respectively. Finally, we test two versions of our own DNN using only either the caption ({\bf DNN (Topic+Caption)}) or the visual information of the image ({\bf DNN (Topic+VGG)}).

Table~\ref{table:results} shows the Top-1 average and nDCG scores obtained. First, we observe that the DNN methods perform better for both the evaluation metrics compared to the baseline methods. They achieve a Top-1 average rating between 1.94 and 2.12 better than the Global PPR, Local PPR, WSABIE, LR and SVM baselines. Specifically, the DNN (Topic+Caption+VGG) method significantly outperforms these models (paired t-test, $p<0.01$). This demonstrates that our simple DNN model captures high-level associations between topics and images. We should also highlight that the network has not seen either the topic or the image during training which is important for a generic model. In the WSABIE model, linear mappings are learned between the text and visual features. This restricts their effectiveness to capture non-linear similarities between the two modalities. 

The DNN (Topic+Caption) model that uses only textual information, obtains a Top-1 Average performance of 1.94. Incorporating visual information (VGG) improves it to 2.12 (DNN (Topic+Caption+VGG)). An interesting finding is that using only the visual information (DNN (Topic+VGG)) achieves better results (2.04) compared to using only text. This demonstrates that images contain less noisy information compared to their captions for this particular task.

\begin{figure}[!t]
\centering
\scriptsize{Topic \#288: surgery, body, medical, medicine, surgical, blood, organ, transplant, health, patient}\\
\subfloat[][\emph{3.0}]
{\includegraphics[width=2cm,height=2cm]{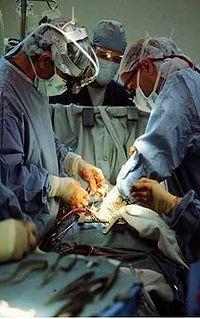}} \quad
\subfloat[][\emph{2.8}]
{\includegraphics[width=2cm,height=2cm]{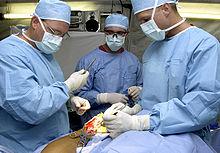}} \quad
\subfloat[][\emph{2.9}]
{\includegraphics[width=2cm,height=2cm]{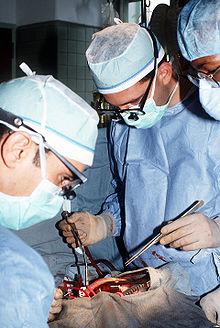}} \\
\scriptsize{Topic \#99: wedding, camera, bride, photographer, rachel, lens, sarah, couple, guest, shot}\\
\subfloat[][\emph{0.4}]
{\includegraphics[width=2cm,height=2cm]{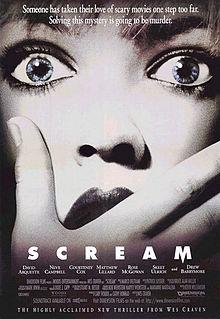}} \quad
\subfloat[][\emph{0.8}]
{\includegraphics[width=2cm,height=2cm]{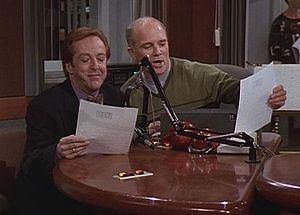}} \quad
\subfloat[][\emph{0.8}]
{\includegraphics[width=2cm,height=2cm]{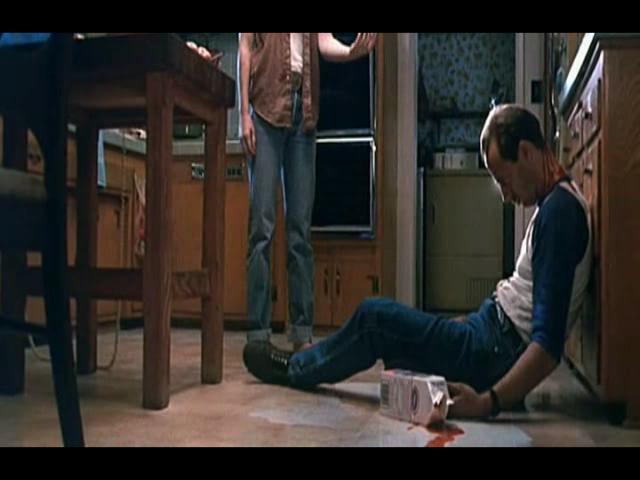}} \\
\caption{A good and a bad example of topics and the top-3 images (left-to-right) selected by the DNN (Topic+Caption+VGG) model from the candidate set. Subcaptions denote average human ratings.}
\label{fig:qualitative}
\end{figure}

The DNN models also provide a better ranking for the image candidates. The nDCG scores for the majority of the DNN methods are higher than the other methods. DNN (Topic+Caption+VGG) consistently obtains the best nDCG scores, 0.79, 0.80 and 0.81 respectively. 
Figure~\ref{fig:qualitative} shows two topics and the top-3 images selected by the DNN (Topic+Caption+VGG) model from the candidate set. The labels selected for the topic \#288 are all very relevant to a \emph{Surgical operation}. On the other hand, the images selected for topic \#99 are irrelevant to \emph{Wedding photography}. For this topic the candidate set of labels do not contain any relevant images.

\section{Conclusion}\label{sec:conclusion}
We presented a deep neural network that jointly models textual and visual information for the task of topic labeling with images. Our model is generic and works for any unseen pair of topic and image. Our evaluation results show that our proposed approach significantly outperforms the state-of-the-art method \cite{Aletras2013a} and a relevant method originally utilized for image annotation \cite{Weston2010}. 

\bibliographystyle{splncs03}
\bibliography{library}

\end{document}